\RequirePackage{fix-cm}
\documentclass[twocolumn,natbib]{svjour3}          
\smartqed  
\usepackage{epsfig}
\usepackage{graphicx}
\usepackage{subfigure}
\usepackage{amsmath}
\usepackage{amssymb}
\usepackage{algorithm, algorithmic}
\usepackage{bbm}
\usepackage[shortcuts]{extdash}

\makeatletter
\renewenvironment{subarray}[2][c]{%
  \if#1c\vcenter\else\vbox\fi\bgroup
  \Let@ \restore@math@cr \default@tag
  \baselineskip\fontdimen10 \scriptfont\tw@
  \advance\baselineskip\fontdimen12 \scriptfont\tw@
  \lineskip\thr@@\fontdimen8 \scriptfont\thr@@
  \lineskiplimit\lineskip
  \ialign\bgroup\ifx c#2\hfil\fi
    $\m@th\scriptstyle##$\hfil\crcr
}{%
  \crcr\egroup\egroup
}
\makeatother
\renewcommand{\substack}[2][c]{\subarray[#1]{c}#2\endsubarray}

\begin{document}

\title{Local Variation as a Statistical Hypothesis Test
}


\author{Michael Baltaxe         \and
        Peter Meer              \and
        Michael Lindenbaum
}


\institute{M. Baltaxe \at
              Orbotech Ltd., Yavne 8110101, Israel. \\
              Tel.: +972-8-942-3879\\
              Fax.: +972-8-943-8769 \\
              \email{michael.baltaxe@orbotech.com}             \\
              This work was conducted while M. Baltaxe was with the Computer Science Department, Technion - Israel Institute of Technology.
           \and
           P. Meer \at
              Department of Electrical and Computer Engineering, Rutgers University, Piscataway, NJ 08854. \\
              Tel.: +1-848-445-5243 \\
              Fax.: +1-848-445-2820 \\
              \email{meer@cronos.rutgers.edu}
           \and
           M. Lindenbaum \at
               Computer Science Department, Technion - Israel Institute of Technology, Haifa 32000, Israel. \\
               Tel.: +972-4-829-4331 \\
               Fax.: +972-4-829-3900 \\
               \email{mic@cs.technion.ac.il}
}


\maketitle

\begin{abstract}
The goal of image oversegmentation is to divide an image into several pieces, each of which should ideally be part of an object. One of the simplest and yet most effective oversegmentation algorithms is known as \emph{local variation} (LV) \citep{FelzSegmentation}.
In this work, we study this algorithm and show that algorithms similar to LV can be devised by applying different statistical models and decisions, thus providing further theoretical justification and a well-founded explanation for the unexpected high performance of the LV approach. 
Some of these algorithms are based on statistics of natural images and on a hypothesis testing decision; we denote these algorithms \emph{probabilistic local variation} (pLV). The best pLV algorithm, which relies on censored estimation, presents state-of-the-art results while keeping the same computational complexity of the LV algorithm.
\keywords{Image segmentation \and Image oversegmentation \and Superpixels \and Grouping}
\end{abstract}

\begin{sloppypar}
\section{Introduction}
\label{sec:Intro}
Image segmentation is the procedure of partitioning an input image into several meaningful pieces or segments, each of which should be semantically complete (i.e., an item or structure by itself).
Oversegmentation is a less demanding type of segmentation. The aim is to group several pixels in an image into a single unit called a superpixel \citep{RenMalik} so that it is fully contained within an object; it represents a fragment of a conceptually meaningful structure.

Oversegmentation is an attractive way to compact an image into a more succinct representation. Thus, it could be used as a preprocessing step to improve the performance of algorithms that deal with higher level computer vision tasks. For example, superpixels have been used for discovering the support of objects \citep{RosenfeldExtractingForegroundMasks}, extracting 3D geometry \citep{SurfaceLayout}, multiclass object segmentation \citep{GouldMultiClassSegmentation}, scene labeling \citep{FarabetSceneLabeling}, objectness measurement in image windows \citep{AlexeObjectness}, scene classification \citep{JunejaBlocks}, floor plan reconstruction \citep{CabralFloorplanReconstruction}, object description \citep{DelaitreSemantics}, and egocentric video summarization \citep{LeeEgocentricVideo}.

One may ask whether specialized oversegmentation processes are needed at all, given that several excellent segmentation methods were recently proposed. Working in the high recall regime, these algorithms could yield excellent oversegmentation accuracy. Unfortunately, however, they are complex and therefore relatively slow. The segmentation approach described in \cite{ArbelaezContourDetection}, for example, combines the gPb edge detector and the oriented watershed transform to give very high precision and recall, but requires over 240 seconds per frame (as reported in \cite{DollarStructuredForest}). The approach of \cite{RenISCRA} uses a large number of classifiers and needs 30 seconds.
Moreover, while the recently introduced edge detector \citep{DollarStructuredForest} is both accurate and very fast (0.08 seconds per frame), it does not provide close edges and consistent segmentation. Using this edge detection for hierarchical multiscale segmentation \citep{ArbelaezMultiscaleCombinatorialGrouping}, indeed achieves state-of-the-art results in terms of precision and recall but requires processing time of 15 seconds per image. Thus, the time required for these methods is often too high for a preprocessing stage, and then specialized oversegmentation algorithms, running in one second or less, are preferred.

One of the many popular approaches to oversegmentation is the mean shift algorithm, which regards the image values as a set of random samples and finds the peaks of the associated PDF, thereby dividing the data into corresponding clusters \citep{MeerMeanShift}.
The watershed method is a morphological approach which interprets the gradient magnitude image as a topographical map. It finds the catchment basins and defines them as segments \citep{WatershedMeyer}.
Turbopixels  is a level set approach, which evolves a set of curves so that they attach to edges in the image and eventually define the borders of superpixels \citep{Turbopixels}.
The SLIC superpixel algorithm is
based on k-means restricted search, and takes into account both spatial and color proximity \citep{SLICSuperpixels}.
The entropy rate superpixels (ERS) \citep{EntropyRateSuperpixels} approach partitions the image by optimizing an objective function that includes the entropy of a random walk on a graph representing the image, and a term balancing the segments' sizes.
A thorough description of common oversegmentation methods can be found in \cite{SLICSuperpixels}.

The local variation (LV) algorithm by \cite{FelzSegmentation} is a widely used, fast and accurate oversegmentation method.
It uses a graph representation of the image to iteratively perform greedy merge decisions by evaluating the evidence for an edge between two segments. Although all the decisions are greedy and local, the authors showed that, in a sense, the final segmentation has desirable global properties. The decision criterion is partially heuristic and yet  the algorithm provides accurate results. Moreover, it is efficient, with complexity $O(n \log n)$.

\begin{figure}[tb]
	\begin{center}
		\includegraphics[width=0.4\textwidth]{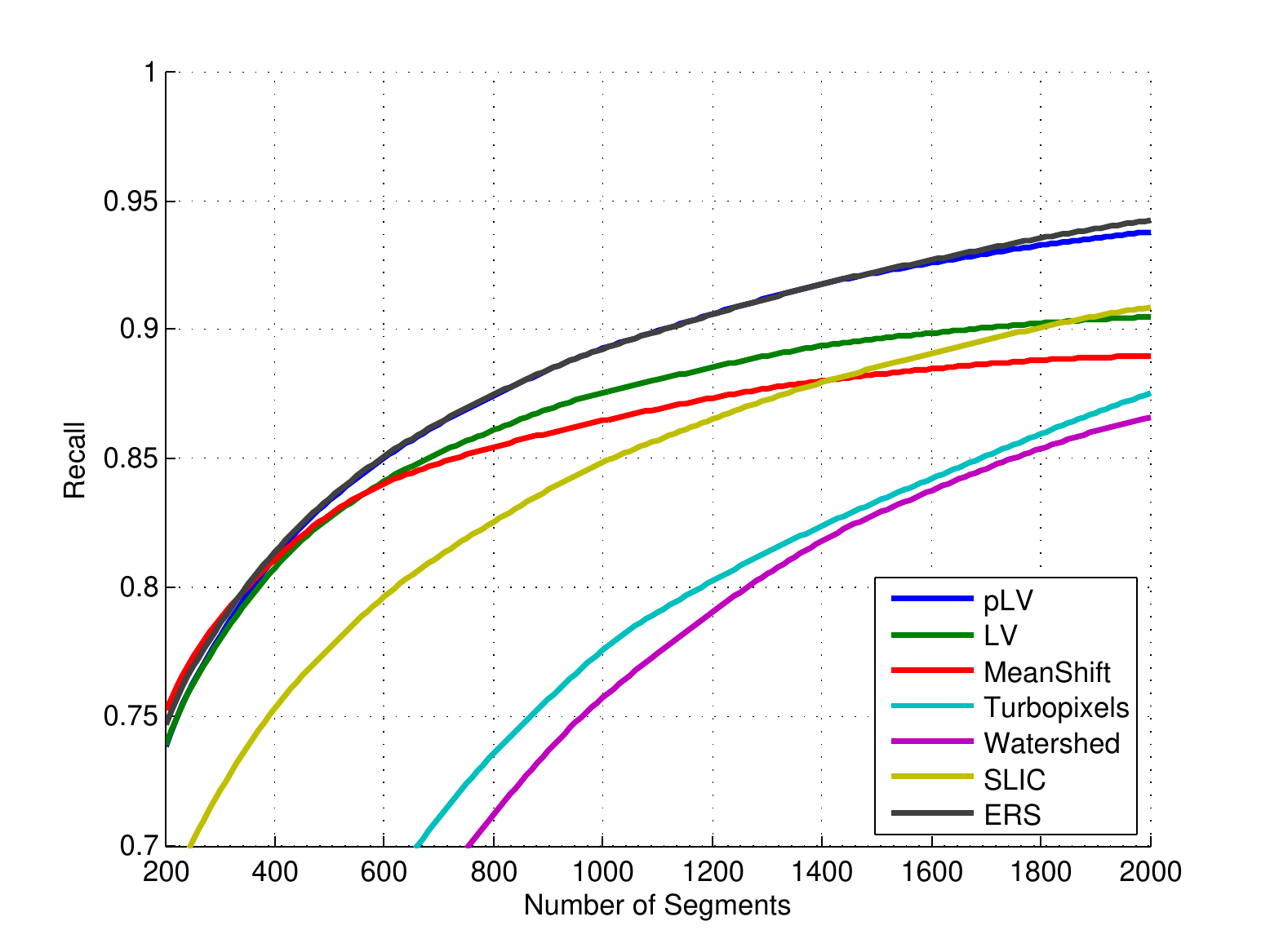}
		\includegraphics[width=0.4\textwidth]{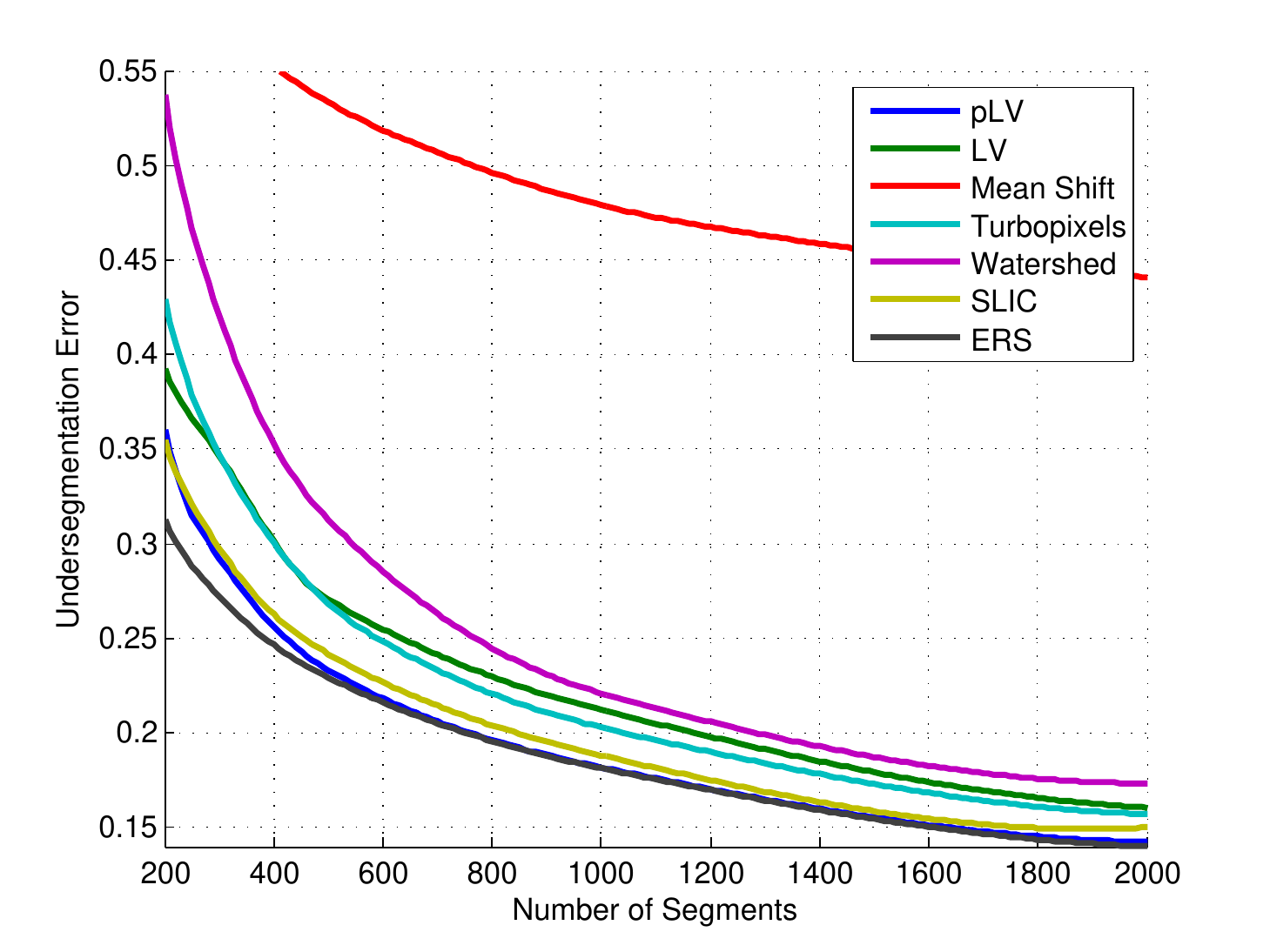}
	\end{center}
	\caption{Recall (top) and undersegmentation error (bottom) for probabilistic local variation and other common algorithms.}
	\label{fig:AllMethodsResults}
\end{figure}

Throughout this paper we use the recall and undersegmentation error 
as measures of quality for oversegmentation, just as is done in \cite{Turbopixels}, \cite{SLICSuperpixels}, and \cite{EntropyRateSuperpixels}.
The recall \citep{MartinLearningNaturalImageBoundaries} is the fraction of ground truth boundary pixels that are matched by the boundaries defined by the algorithm.
The undersegmentation error \citep{Turbopixels} quantifies the area of regions added to segments due to incorrect merges.
Figure \ref{fig:AllMethodsResults} presents a comparison of the aforementioned algorithms, together with probabilistic local variation (pLV), the method introduced in this paper\footnote{Code available at \\http://cis.cs.technion.ac.il/index.php/projects/probabilistic-local-variation}.

Understanding the remarkably good performance of the greedy LV algorithm was the motivation for our work, which makes three contributions:
%
\begin{enumerate}
	\item We briefly analyze the properties of the LV algorithm and show that the combination of all its ingredients is essential for high performance.
	\item We examine several probabilistic models and show that LV-like algorithms could be derived by statistical  consideration. One variation involves the estimation of the maximum sample drawn from a uniform distribution. The second uses natural image statistics and a hypothesis testing decision.
This gives additional viewpoints, further justifies the original version of the algorithm, and explains the excellent performance obtained from such a simple, fast method.
	\item Following these models, we provide a new algorithm, which is similar to LV but relies on statistical arguments. This algorithm meets the best performance among the methods in the literature, at least for oversegmentation to many segments, but is faster.
\end{enumerate}

The structure of the paper is as follows.
We present the LV algorithm in section \ref{sec:LocalVariation}. We study this algorithm empirically in section \ref{sec:EmpiricalTests}. In section \ref{sec:StatisticalInterpretation} we introduce the first LV-like algorithm (based on maximum estimation), which does not perform as well as the original LV, but motivates the alternative and more successful algorithm of section \ref{sec:LocalVariationAsAnHypothesisTestingProblem} (based on natural image statistics). Section \ref{sec:Results} presents our experimental results. A brief discussion of single linkage algorithms (which include the LV and pLV algorithms) as compared to non-single-linkage, hierarchical algorithms is given in section \ref{sec:Discussion}. Finally, section \ref{sec:Conclusion} concludes.

\section{Local Variation Algorithm}
\label{sec:LocalVariation}
The LV algorithm is a single-linkage, graph based,  hierarchical clustering process.
Let $G = (V,E)$ be a graph created from the input image, with vertices on the pixels and edges between each pair of neighboring pixels. Define a weight function on the edges, $w: E \rightarrow \Re$, representing pixel dissimilarity (for example, the RGB color distance between pixels). Likewise, let a component (segment) $C_i$ be a set of connected pixels. The components change throughout the segmentation process, and, initially, the set of components $\{C_i\}$ is the set of pixels.

Following \cite{FelzSegmentation}, define the internal dissimilarity of component $C_i$, denoted by $Int(C_i)$, and a threshold function called the minimum internal difference and denoted $MInt(C_i, C_j)$, as:
\begin{align}
	&Int(C_i) = \underset{e \in MST(Ci)}{\max}w(e)\\
	&MInt(C_i, C_j) = \underset{x \in \{i,j\}}{\min}\left( Int(C_x) + T(C_x) \right),
	\label{eq:FHInternalDifference}
\end{align}
where $MST(C_i)$ is a minimum spanning tree of $C_i$, and $T(C_i)= K/\left| C_i \right|$ is a component dependent function,
in which $K$ is a user controlled parameter and $|C_i|$ denotes the number of vertices in component $C_i$.

\begin{algorithm}[tb]
	\caption{Local Variation Algorithm}
	\label{algo:LocalVariation}
	\textbf{Input:} Weighted graph $G = (V,E)$ with weights $w(e)$, $e \in E$, defined by an image.\\
	\textbf{Output:} Set of components $C_1,...,C_n$ defining segments
	\begin{algorithmic}[1]
		\STATE Sort $E$ by non-decreasing edge weight $\left(e_1, e_2,..., e_m\right)$
		\STATE Initialize segmentation $S^0$ with each vertex being a component
		\FORALL{$q = 1,...,m$}
			\STATE $e_q = (v_i, v_j) \gets$ edge with the $q$th lightest weight
			\STATE $C_i^{q-1} \gets$ component of $S^{q-1}$ containing $v_i$
			\STATE $C_j^{q-1} \gets $ component of $S^{q-1}$ containing $v_j$
			\IF{$\left( w(e_q) \leq MInt \left( C_i^{q-1}, C_j^{q-1} \right) \right) \wedge (C_i^{q-1} \neq C_j^{q-1})$}
				\STATE $S^q = S^{q-1} \cup \left\{ C_i^{q-1} \cup C_j^{q-1} \right\} \setminus \left\{ C_i^{q-1}, C_j^{q-1} \right\}$
			\ELSE
				\STATE $S^q = S^{q-1}$
			\ENDIF
		\ENDFOR
		\STATE \textbf{Postprocessing:} Merge all small segments to the neighbor with closest color.
	\end{algorithmic}
\end{algorithm}

The LV algorithm for image oversegmentation is presented in algorithm \ref{algo:LocalVariation}.
Intuitively, we can see that two components are merged only if the lightest edge that connects them is lighter than the heaviest edge in the MST of the components plus a margin. Since the edges are sorted in step 1, the edges causing merges are exactly those that would be selected by Kruskal's MST algorithm \citep{KruskalAlgorithm}.
The parameter $K$ controls the number of segments in the output segmentation: increasing $K$ implies that more edges satisfy the merge condition and more merges are performed.


Oversegmentation algorithms usually include a post-processing stage where small segments are removed (line 13 in algorithm \ref{algo:LocalVariation}). We consider a segment as small when its size is 10\% of the average expected segment size, and merge it to its neighbor with the smallest color difference.

%
Compared to other oversegmentation algorithms, LV is among the best in terms of recall and running time. It is thus often the method of choice even though its undersegmentation error is not as small as that of some other algorithms; see \cite{SLICSuperpixels} and figure \ref{fig:AllMethodsResults}.


The high accuracy obtained by the greedy LV algorithm is impressive. It is mainly due to the adaptive threshold $Int(C_i) + T(C_i)$, which depends on two components: the distribution of weights within the segments and their size. The particular combination of these two components meets the criterion used to decide whether a segmentation is too fine / coarse in \cite{FelzSegmentation}, but is not theoretically supported otherwise. In this paper we analyze the LV algorithm empirically and propose two statistical interpretations that lead, eventually, to LV-like algorithms, that follow a statistical decision procedure. One of these new algorithms, denoted as probabilistic local variation (or more specifically, pLV-ML-Cen), maintains LV's desirable properties, and improves its recall and undersegmentation error.

\section{Empirical Study of the LV Algorithm}
\label{sec:EmpiricalTests}
%
To analyze the importance of each of the two aforementioned components in LV, we test reduced versions of the algorithm by systematically removing each component:


\noindent {\bf 1. Greedy Merging:}
By setting $MInt(C_i,C_j) = \infty$ the algorithm depends neither on the distribution nor on the segment sizes. This implies that the segments are merged greedily in non-decreasing edge weight.
%

\noindent {\bf 2. LV with a Constant Threshold:}
By setting $T(C_i) = K$, where $K$ is a constant, makes the decision size independent (but distribution dependent).
Performance is substantially reduced and very large segments are created.
\noindent {\bf 3. Area Based Merging:}
The condition in line 7 of algorithm \ref{algo:LocalVariation} is replaced by $\underset{x \in \{i,j\}}{\min} \left| C_{x}^{q-1} \right| < K$, where $K$ is a constant. This condition depends only on the segment size but not on the distribution, yielding superpixels with roughly the same area.

\noindent {\bf 4. LV without Removing Small Segments:}
A post-processing step in LV removes small components (line 13 in algorithm \ref{algo:LocalVariation}).
Without this step, a lot of very small, meaningless segments remain, implying that many erroneous merges are performed to obtain a prespecified number of segments.

\begin{figure}[tb]
	\begin{center}
		\includegraphics[width=0.4\textwidth]{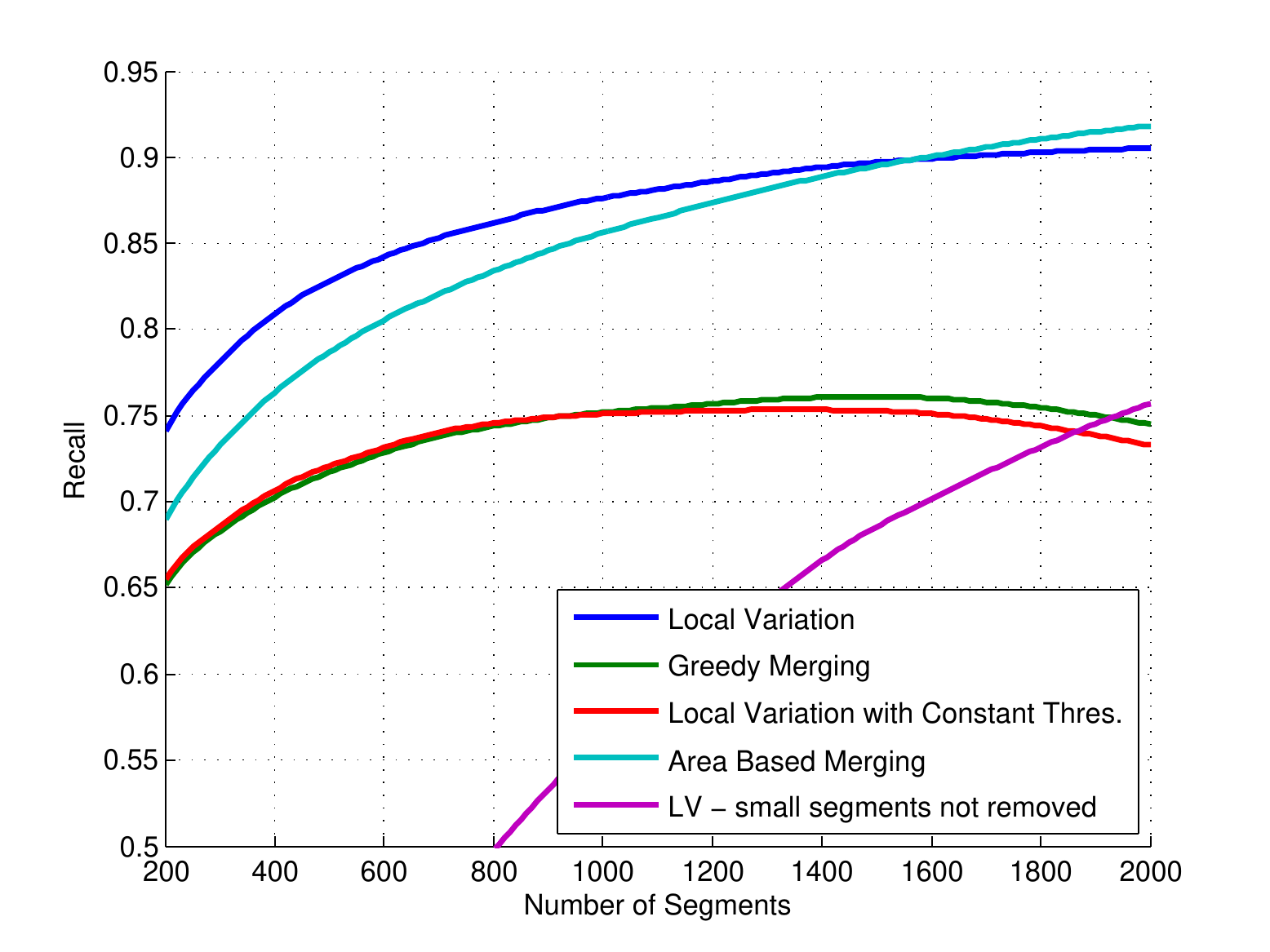}
	\end{center}
	\caption{The LV method against its reduced versions. No reduced version could achieve the performance of the original algorithm.}
	\label{fig:FHExperimentsRecallResults}
\end{figure}

Figure \ref{fig:FHExperimentsRecallResults} compares the recall of LV and its reduced versions. All reduced versions yield lower recall than the original algorithm. Thus we conclude that both the distribution and size dependent terms are crucial to LV's performance.

\section{Interpreting LV as Maximum Estimation}
\label{sec:StatisticalInterpretation}
\subsection{Estimating the Maximum Sample Drawn from a Uniform Distribution}
\label{EstimatingTheMaximumOfAUniformDistribution}
The threshold used in LV has several possible statistical interpretations. In this section we consider one interpretation, which suggests that LV's decision rule is similar to maximum value estimation.

Consider a set of samples drawn from a uniform distribution specified by an unknown interval $\left[min_U \!, max_U \right]$. We want to estimate the parameters of the uniform distribution, $min_U, max_U$, from the samples\footnote{This is the continuous version of the problem known in the statistical theory literature as the German tank problem \citep{GermanTankProblem}, because its solution was used by the Allies in WW2 to estimate the number of tanks produced by the Germans from the serial numbers of captured tanks.}.
For the special case where $min_U=0$, let $m$ be the sample maximum and $S$ be the set size. Then, the minimum variance unbiased estimator for the maximum value is given by \citep{StatisticsLarsenMarx}
\begin{equation}
	\hat{max}_U = m + \frac{m}{S}.
	\label{eq:MaximumSampleEstimator}
\end{equation}

\subsection{Interpreting LV as Maximum Estimation}
\label{sec:InterpretationOnLocalVariation}
The estimate \eqref{eq:MaximumSampleEstimator} seems similar to the threshold expression used in the local variation algorithm (algorithm \ref{algo:LocalVariation} and \cite{FelzSegmentation}). Both expressions contain two terms. The first is a distribution related term, $m$, which is the maximal observed value. The second term in both cases is size dependent. The two expressions differ in that, in expresion \eqref{eq:MaximumSampleEstimator}, the size dependent term depends also on the maximal observed value, $m$.
Thus, we hypothesize that an LV-like process would be obtained by considering the weight values in each segment, estimating their maximum under the uniform distribution assumption, and testing whether a new weight falls below the estimated maximum and therefore belongs to the same distribution. If the weight satisfies this test for the two segments, then merging them is justified.

Thus the algorithm is exactly like algorithm \ref{algo:LocalVariation} (LV), except that $MInt(C_i,C_j)$ is replaced by:
\begin{equation}
	\label{eq:LVMaxU}
	MInt(C_i,C_j) = \underset{x \in \{i,j\}}{\min}\left( Int(C_x) + \frac{Int(C_x)}{|C_x|} + c \right)\!.
\end{equation}

In theory the constant $c$ should be $0$. However, due to image quantization, the maximum observed value in a small segment is often $0$, which prevents any further merges. Note also that the difference between two continuous grey levels may be almost 1, and yet, the difference between their quantized values is zero. To allow for these errors, we set $c = 1$.
Otherwise, this LV-like process \eqref{eq:LVMaxU}, denoted as LV-MaxEst, is parameterless.

Figure \ref{fig:FHTests} presents an example. 
Note that the overall segmentation still does not seem natural: the segments on the bird look natural, but those in the sky do not.

\subsection{A Controllable Segmentation Based on Maximum Estimation}
\label{sec:UsingTheMaximumSampleStatistics}
Being parameterless, the decision rule based on the maximum estimation does not allow us to control the oversegmentation level. A simple, controllable extension would be to change $MInt(C_i,C_j)$ to:

\begin{equation}
\label{eq:MaximumSampleStatisticsK}
	MInt(C_i,C_j) = \underset{x \in \{i,j\}}{\min}\left( Int(C_x) + \frac{k \cdot Int(C_x)}{|C_x|} + c \right),
\end{equation}
where the constant $k$ controls the number of superpixels. We denote this method LV-MaxEst-c (where the suffix ``c'' stands for ``controlled'').

Interpreting LV as a maximum estimation problem sheds some light on the algorithm; however, some of the assumptions used in this section are not accurate and the decision rule in equation (\ref{eq:MaximumSampleStatisticsK}) actually does not perform as well as the original LV algorithm (see section \ref{sec:Results}). This motivates the alternative model presented next.

\section{Interpreting LV as Hypothesis Testing}
\label{sec:LocalVariationAsAnHypothesisTestingProblem}
The statistical interpretation in section \ref{sec:StatisticalInterpretation} is closely related to the original LV formulation but the assumption of a uniform distribution for edge weights seems unjustified. In this section, we present an alternative statistical model associated with natural image statistics.

\subsection{Natural Image Statistics}
\label{sec:NaturalImageStatistics}
Natural image statistics have been intensively explored over the last two decades. Statistical models consider specific image descriptors such as wavelet coefficients \citep{NonParametricMultiScaleStatisticsModel} or intensity difference between adjacent pixels \citep{ProbabilityModelsForClutterInNaturalImages} and characterize them statistically.
A common way to model the behavior of these image descriptors is by means of the generalized Gaussian distribution \citep{MallatWavelets},
\begin{equation}
	P(x) = \frac{\beta}{2\alpha\Gamma(1/\beta)}e^{-\left(x / \alpha\right)^\beta},
\end{equation}
where, typically, $\alpha$ varies depending on the descriptor used and  $\beta$ falls in the range $[0.5,0.8]$ \citep{SrivastavaNaturalImageStats}. This model has been successfully used for, e.g., image denoising \citep{DenoisingWithGeneralizedLaplacian} and image segmentation \citep{NaturalImageStatsForSegmentation}.

We are interested in the weights of the graph edges, which are either the absolute differences between LUV color vectors or simply intensity differences. The intensity differences are closely related to some wavelet coefficient and to gradient strengths, both of which were modeled with the generalized Gaussian distribution \citep{MallatWavelets,GeneralizeGausianGradients}.

\begin{figure}[tb]
	\begin{center}
		\raisebox{-0.5\height}{\includegraphics[width=.19\textwidth]{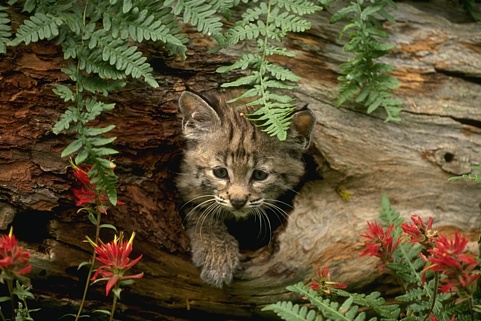}}
		\raisebox{-0.5\height}{\includegraphics[width=.282\textwidth]{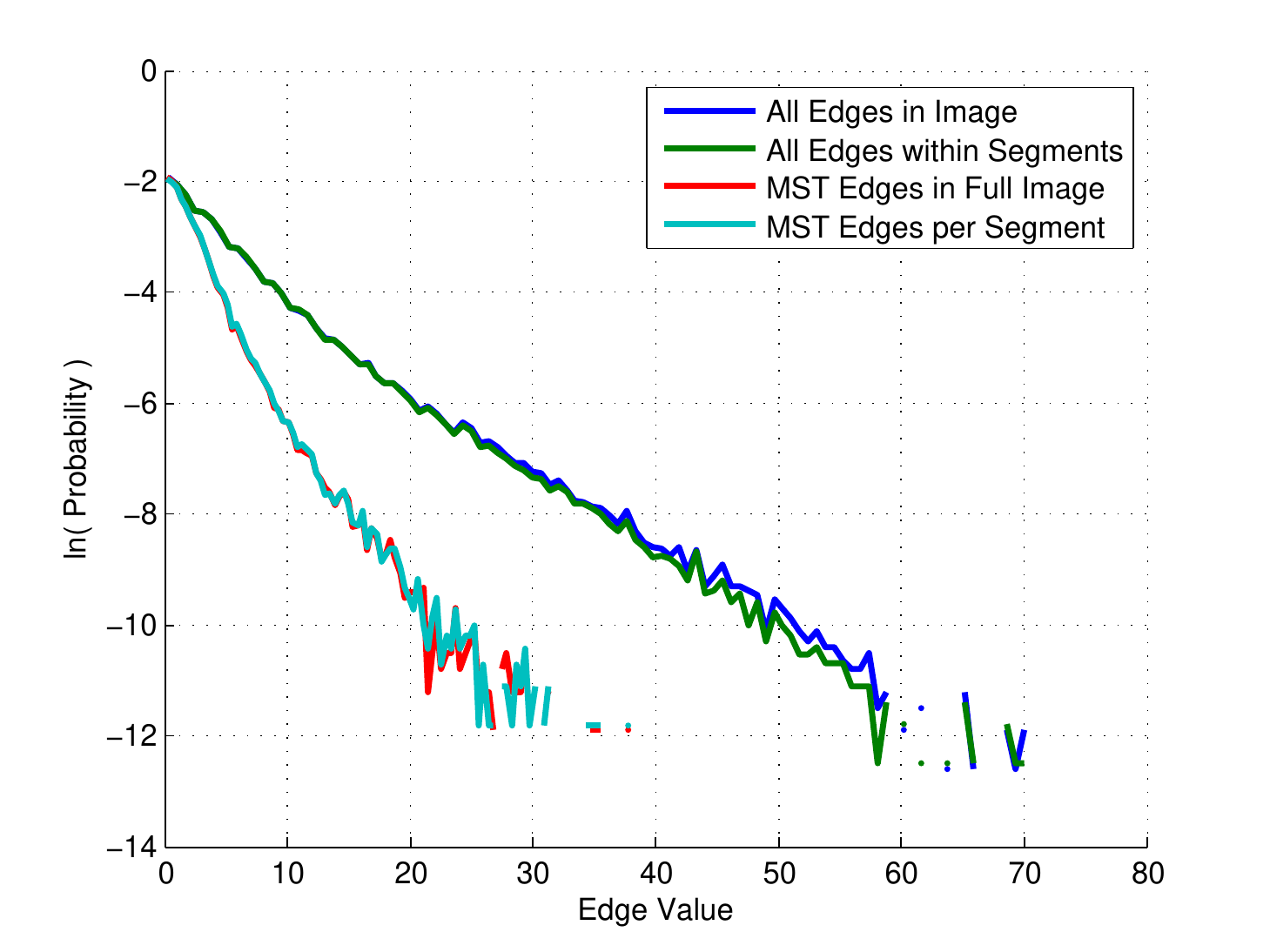}}
	\end{center}
	\caption{A natural image and semilog plot with the edge weight histogram of several edge sets. See text for details.}
	\label{fig:HistogramsOfEdges}
\end{figure}

The population we consider is somewhat different, however. We are interested only in weights that are part of the MST and are inside segments (and not between them). We checked the validity of the exponential model, a particular case of the generalized Gaussian distribution with $\beta = 1$, on several images and found that the exponential assumption is reasonable; see figure \ref{fig:HistogramsOfEdges}, which shows one image example and 4 plots of the LUV edge weight statistics:
\begin{enumerate}
	\item Weights of all edges in the image.
	\item Weights of all edges within image segments, as marked by a human (as given in BSDS300 \citep{BerkeleyDataset}).
	\item Weights of edges in an MST of the full image.
	\item Weights of edges in an MST of each segment, as marked by a human.
\end{enumerate}

As expected, the distributions of the edge weights in the MSTs are biased towards lower weights compared to the distributions obtained from the edges in the complete image. On the other hand, the edge weight distributions within the segments and throughout the entire image are similar. This is also expected because the number of edges crossing segments is relatively small. Clearly, all distributions are close to an exponential law (observed as a straight line in the semilog plot). In this work, we shall therefore assume that the model underlying the distribution of edge weights is the exponential distribution:
\begin{equation}
	\label{eq:exponentialDistribution}
	P(x) = \lambda e^{-\lambda x}.
\end{equation}

\subsection{Merge Decisions from Hypothesis Testing}
\label{sec:HypothesisTesting}
Consider the local merging context where we need to decide whether two segments $S_i,S_j$ merge along edge $e$.
We propose to make this decision by testing the hypothesis that the weight $w(e)$ of the edge $e$ belongs to the distribution $P_i(x) = \lambda_i e^{-\lambda_i x}$ of the weights in each one of the segments. If this hypothesis is rejected for at least one of the segments, the segments are not merged. Otherwise the segments are merged. We refer to this method as \emph{probabilistic local variation (pLV)}.

To test the hypothesis that $w(e)$ belongs to $S_i$, we consider the probability:
\begin{equation}
	\label{eq:HypothesisProbability}
	P_i(x > w(e)) = \int^\infty_{w(e)} \lambda_i e^{-\lambda_i x} dx = e^{-\lambda_i w(e)}.
\end{equation}
The hypothesis is rejected with a level of significance $\delta$ whenever $P_i(x > w(e)) < \delta$.

Thus, the probabilistic local variation approach uses the following alternative rule for deciding whether two segments should be merged:
\begin{enumerate}
	\item Let $e^*$ be the edge with the minimum weight connecting two segments, $S_i, S_j$.
	\item For each segment $S_a \in \{S_i, S_j\}$, fit an exponential distribution, $P_a(x)$, to the weights in it.
	\item For each segment $S_a \in \{S_i, S_j\}$, test the hypothesis that $e^*$ belongs to the corresponding distribution using the hypothesis test, $P_a(x > w(e^*)) < \delta$.
	\item If the hypothesis is rejected in at least one of the tests, do not merge. Otherwise merge.
\end{enumerate}

In the rest of this section we describe several ways to estimate the parameters $\lambda_i$ and the implied distribution.

A straightforward estimate for $\lambda$ would be the maximum likelihood (ML) estimator. Given a population of $n$ samples, $\left\{ x_1, x_2, ..., x_n \right\}$, drawn i.i.d.
from the exponential distribution \eqref{eq:exponentialDistribution}, then the maximum likelihood estimator for the parameter, $\hat{\lambda}_{ML}$, is
\begin{equation}
	\label{eq:lambdaML}
	\hat{\lambda}_{ML} = \frac{1}{\bar{x}} = \frac{n}{\sum_{i = 1}^{n}{x_i}}.
\end{equation}
When using this estimator, we will denote the probabilistic local variation as pLV-ML.

The ML estimator is, however, noise prone and highly unstable when the sample size is small, as is the case when the merged segments are small. One way to make a robust decision is by using confidence intervals (CI) for $\hat{\lambda}_{ML}$. The symmetric $100(1-\alpha)$\% CI for $\hat{\lambda}_{ML}$, for a population of $n$ samples drawn from the exponential distribution, is given by:
\begin{equation}
  \label{eq:ConfidenceIntervalLambda}
	\left( \frac{\chi^2_{1-\alpha/2,2n}}{2n}\hat{\lambda}_{ML}, \frac{\chi^2_{\alpha/2,2n}}{2n}\hat{\lambda}_{ML} \right),
\end{equation}
where $\chi^2_{p,\nu}$ is a value specifying the tail of weight $p$ in a $\chi^2$ distribution with $\nu$ degrees of freedom
(see section 7.6 of \cite{Ross2009Probability} for details regarding parameter estimation for the exponential distribution).
$\chi^2_{p,\nu}$ increases approximately linearly with $\nu=2n$, and for large $n$, converges to $\nu=2n$, implying that the
confidence interval decreases with the segment size, as intuitively expected.

The question of which value should be chosen within the CI remains. The effect of choosing a particular $\lambda$ value is not straightforward because changes in $\lambda$ influence other parameters. Suppose we are interested in a final segmentation with $N$ superpixels. Choosing $\lambda$, say, at the lower limit of the confidence interval, must be compensated for by increasing $\delta$. Otherwise, the threshold will be higher and too many merges will be performed. As explained in section \ref{sec:ParameterEstimationUnderBiasSampling}, the edge weights used for estimating $\lambda$ are biased towards smaller values, which makes $\hat{\lambda}_{ML}$ biased to larger values. Therefore, we prefer the lower limit of the CI \eqref{eq:ConfidenceIntervalLambda}, and reject a merge whenever
	$P(x > w(e)) = e^{-w(e) \hat{\lambda}_{ML} \frac{\chi^2_{1-\alpha/2,2n}}{2n}} < \delta$,
or equivalently, whenever
\begin{equation}
	\label{eq:RejectionRule}
	w(e) > \ln \left( 1/\delta \right) / \left( \hat{\lambda}_{ML} \frac{\chi^2_{1-\alpha/2,2n}}{2n} \right).
\end{equation}

Replacing $\hat{\lambda}_{ML}$ with the lower limit of the CI for $\lambda$ (method denoted as pLV-ML-CI) indeed yields better recall, very similar to that of LV; see figure \ref{fig:HypothesisTestResults} (top).
The role of the confidence parameter $\delta$ is analogous to that of $K$ in the original LV algorithm: making $\delta$ larger results in more segments.

Note also that qualitatively, the algorithm behaves according to the LV principle and gives priority to the merging of smaller superpixels. This is not immediately clear, however, because, as discussed above, choosing $\lambda$ at the lower limit of CI
is compensated for by increasing $\delta$. This compensation, however, is non-uniform. While the level of significance $\delta$ is uniform and applies to all merging decisions, smaller segments are associated with larger confidence intervals, which makes the lower limit of their confidence intervals lower on average. Therefore, the threshold on the edge weight \eqref{eq:RejectionRule} is higher for smaller superpixels, which gives them priority to merge.

While this model achieves nice segmentations and sheds some new light on the LV algorithm, we observe that the statistical assumption underlying the estimation of $\lambda$ -- that the measurements are obtained by i.i.d. sampling -- seems inaccurate.  First, the samples we have are sampled in nearby locations, which makes them correlated. Moreover, they are sampled with preference for lower weight edges. The first problem seems to be minor because the weights are derivative values, which are less correlated than the image intensities. The second problem, preferring smaller values, is considered in the next subsection.

\subsection{Parameter Estimation under Biased Sampling}
\label{sec:ParameterEstimationUnderBiasSampling}
Contrary to the assumption used in section \ref{sec:HypothesisTesting}, the weights are sampled from the lightest to the heaviest and are not drawn i.i.d. Thus, at each step the population inside a segment is biased towards low values. In what follows we aim to correct this bias.

Suppose the segment has $m+1$ pixels, and at some point in time our method of sampling provides us with $n$ measurements, which are the lowest $n$ elements in the MST of the segment, which contains $m$ edges. Our task is to estimate the value of $\lambda$. 
%
The structure of this problem is an approximation to the one defined by \emph{type II censoring}, in which $m$ random variables are drawn i.i.d. but only the smallest $n<m$ values are observed (an approximation since the true segment may be split between several segments specified by the algorithm). This type of sampling is common in reliability estimation, where one tries to study the failure rate of some process/machine by analyzing only a subset composed of the samples which failed first \citep{EpsteinLifeTesting}. Under these conditions, and following \citeauthor{EpsteinLifeTesting}, the maximum likelihood for $\lambda$ can be derived as follows:

A sequence of size $m$ is called partially $n$-ordered if the first $n$ elements in it are non-decreasing (or non-increasing), and not larger (or not smaller) than any
of the remaining $m-n$ elements. Consider a sequence, $Y$, of $m$ samples drawn i.i.d. from some probability distribution. A partially $n$-ordered sequence, $X$, may be generated from it by choosing the $n$ smallest elements from $Y$, and making them the first $n$ elements of $X$, in non-decreasing order. The remaining $m-n$ elements in $X$ are identical to the remaining $m-n$ elements in $Y$. Their order is the same order as in $Y$.

%

Note that many sequences $Y$ may correspond to the same partially $n$-ordered $X$. The first $n$ elements may come from $H = \frac{m!}{(m-n)!}$ different sets of locations, implying that the probability of getting a particular partially $n$-ordered sequence is $H$ times the probability of getting the original sequence $Y$ of i.i.d. drawn samples. Therefore, the joint probability density of observing the values $x_1,...,x_n$ is
\begin{align}
	f&(x_1,...,x_n)\! =\! H P(x_1,...,x_n) P(x_{n+1} > x_n,..., x_m > x_n) \notag \\
	    &= H \cdot P(x_1)\cdots P(x_n) \cdot P(x_{n+1} > x_n)\cdots P(x_m > x_n) \notag \\
	    &= H \cdot P(x_1)\cdots P(x_n) \cdot \left(\int^\infty_{x_n} \lambda e^{-\lambda x} dx\right)^{m-n} \notag \\
	    &= H \cdot \lambda e^{-\lambda x_1} \cdots \lambda e^{-\lambda x_n} \cdot \left( e^{-\lambda x_n} \right)^{m-n} \notag \\
	    &= H \lambda^n \cdot e^{-\lambda \sum^n_{i=1} x_i} \cdot e^{-\lambda (m-n) x_n} \notag \\
	    &= H \lambda^n \cdot e^{-\lambda \left( \sum^n_{i=1} x_i + (m-n) x_n \right)}.
\end{align}
The value of $\lambda$ yielding the maximum likelihood for this function can be found by differentiating $\ln f(x_1,...,x_n)$ with respect to $\lambda$ and equating to zero, yielding
%
\begin{equation}
	\label{eq:EstimationLambda}
	\hat{\lambda}_{ML-Cen} = \frac{n}{\sum^n_{i=1} x_i + (m-n) x_n},
\end{equation}
where the notation ML-Cen is for \emph{censored maximum likelihood}.

Finally, consider a segment $C$ with edge weights being $\{x_1,...,x_n,...,x_m\}$, and the merging decision when the edge under study is $e$. Then, combining equations \eqref{eq:RejectionRule} and \eqref{eq:EstimationLambda}, the one-sided condition for merge rejection becomes:
\begin{align}
	\label{eq:RejectionRuleExpanded}
	w(e) &> HypThr(C) \notag \\
	     &= \left. \frac{2 \ln \left( 1/\delta \right) \left( \sum^n_{i=1} x_i + (m-n) x_n \right)}{\chi^2_{1-\alpha/2,2n}} \right|_{\substack[b]{x_i \in C \\ n = \left| C \right|}},
\end{align}
where $m$ is a user specified parameter reflecting the expected size of the true segment. In the case that $m=n$, $\hat{\lambda}_{ML-Cen}$ converges to $\hat{\lambda}_{ML}$, which we use also for $n > m$.
To gain further intuition about this threshold, recall that $\chi^2_{1-\alpha/2,2n}$ grows almost linearly with the segment size $n$.

In summary, our hypothesis testing algorithm for image oversegmentation (denoted pLV-ML-Cen) is exactly as algorithm \ref{algo:LocalVariation}, but the minimum internal difference between two components is set to:
\begin{equation}
	MInt(C_i, C_j) = \underset{x \in \{i,j\}}{\min} HypThr(C_x).
\end{equation}

\begin{figure}[tb]
	\begin{center}
		\includegraphics[width=.15\textwidth]{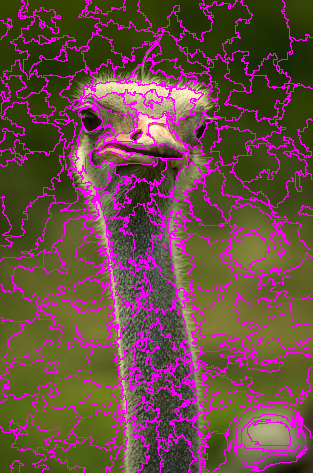}
		\includegraphics[width=.15\textwidth]{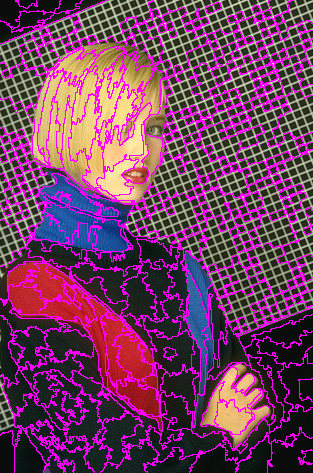}
		\includegraphics[width=.15\textwidth]{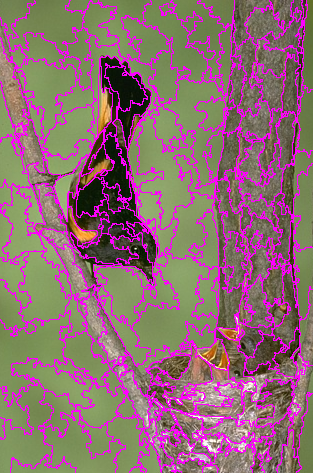} \\
		\includegraphics[width=.15\textwidth]{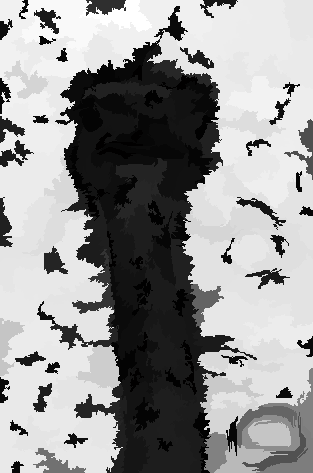}
		\includegraphics[width=.15\textwidth]{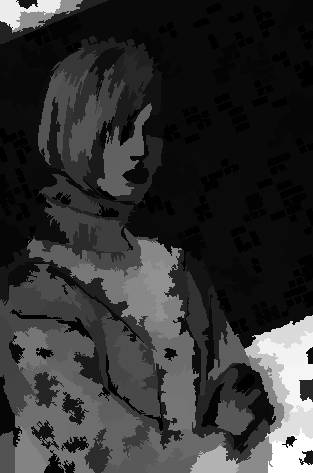}
		\includegraphics[width=.15\textwidth]{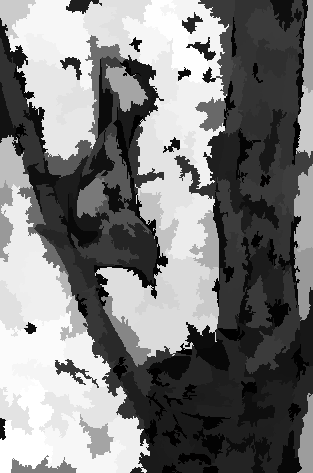}
	\end{center}
	\caption{Bottom: visualization of $\hat{\lambda}$ for each segment. The brighter the segment, the larger the value of $\hat{\lambda}$. Thus, white represents smooth segments and black represents textured ones. Top: corresponding oversegmentation.}
	\label{fig:VisualizationLambda}
\end{figure}

\begin{figure}[tb]
	\begin{center}
		\includegraphics[width=.11\textwidth]{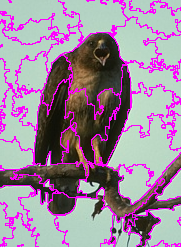}
		\includegraphics[width=.11\textwidth]{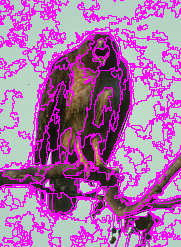}
		\includegraphics[width=.11\textwidth]{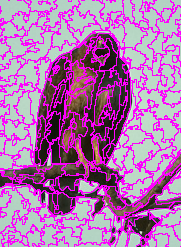}
		\includegraphics[width=.11\textwidth]{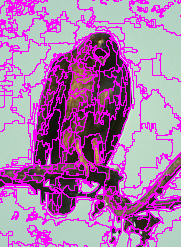}
	\end{center}
	\caption{Segmentations obtained by the methods described in the paper (left to right): LV-MaxEst, LV-MaxEst-c, pLV-ML-Cen and LV. The last 3 methods are tuned to have roughly the same number of segments.}
	\label{fig:FHTests}
\end{figure}

The value of $\lambda$ for a given segment characterizes the distribution of its edges. The larger the variability within a segment, the smaller the value of $\lambda$; see figure \ref{fig:VisualizationLambda}.
Figure \ref{fig:FHTests} presents an example segmentation of the probabilistic local variation method with correction for biased sampling. Note that probabilistic local variation appears to have more segments than the original LV. This happens because the LV algorithm produces many elongated and hardly visible segments along the boundaries.

Probabilistic local variation with censoring is based on different principles and yet follows the same basic behavior as the LV algorithm. We consider the decision condition \eqref{eq:RejectionRuleExpanded} and observe the following four properties:
\begin{enumerate}
	\item As discussed above, the denominator $\chi^2_{1-\alpha/2,2n}$ grows with the size of the segment, $n$, while the numerator decreases with $n$. Thus, the threshold decreases with $n$ and gives preference to merges of small segments.
	
	\item The heaviest weight in the MST, $x_n$, appears in the numerator of $HypThr(C)$. Thus, heavier $x_n$ leads to a higher threshold and to a more likely merge decision.
	
	\item For small segments, where $n \ll m$, the importance of the heaviest edge, $x_n$, is amplified by a factor linear in $(m-n)/n$. For larger segments, though, the amplification factor is smaller, making the average weight of the segment edges more important.

	\item It is straightforward to show that there is a predicate for which probabilistic LV leads to segmentation which is not too fine in the sense specified in \cite{FelzSegmentation}. Showing that it is not too coarse \citep{FelzSegmentation} seems harder because the sampling order is not tightly related to our threshold criterion.
\end{enumerate}
%

\subsection{Complexity of Hypothesis Testing}
\label{sec:ComputationalComplexityOfHypothesisTesting}
To calculate the threshold \eqref{eq:RejectionRuleExpanded}, we need to keep three values for each segment: the sum of its elements, the value of the last (heaviest) edge added to it, and its number of elements. All these values are updated in $O(1)$, and thus the complexity of the hypothesis testing method is exactly the same as that of the LV algorithm, namely $O(n \log n)$.

\section{Experimental Results}
\label{sec:Results}
\label{sec:OversegmentationQualityMeasures}

\subsection{Testing Probabilistic Local Variation}
\label{sec:TestingProbabilisticVersionsOfTheLocalVariation}
Following common practice, we use the boundary recall and the undersegmentation error as quantitative performance measures.
The recall \citep{MartinLearningNaturalImageBoundaries} is the fraction of ground truth boundary pixels that are matched by the boundaries defined by the algorithm.
%
The undersegmentation error \citep{Turbopixels} measures the area of incorrect merges of true segments (or parts of them); see \cite{MartinLearningNaturalImageBoundaries,Turbopixels} for implementation details.
%
All experiments were performed on
BSDS300 (test) \citep{BerkeleyDataset}.

\begin{figure}[tb]
	\begin{center}
		\subfigure{\includegraphics[width=0.4\textwidth]{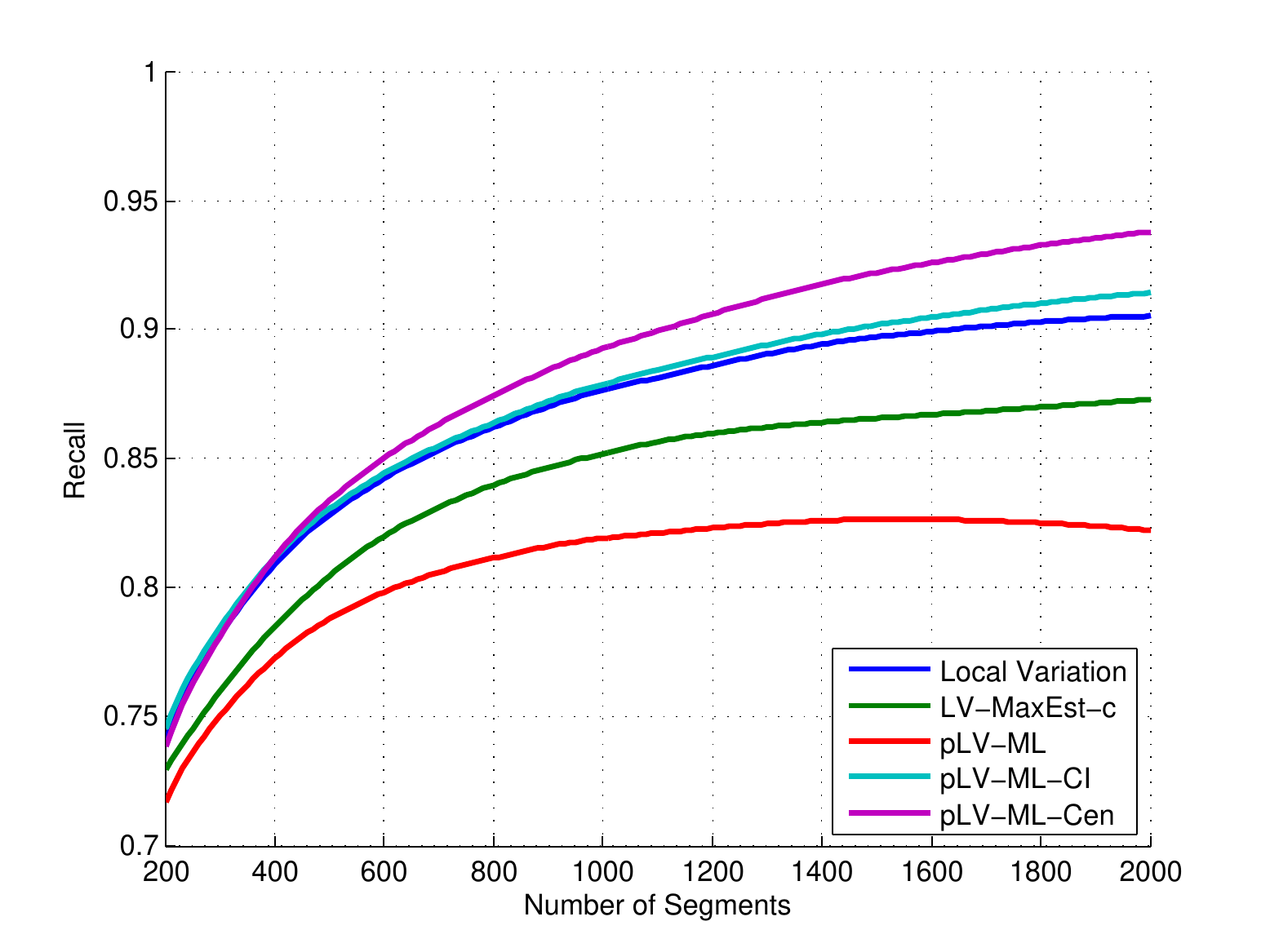}}
		\subfigure{\includegraphics[width=0.4\textwidth]{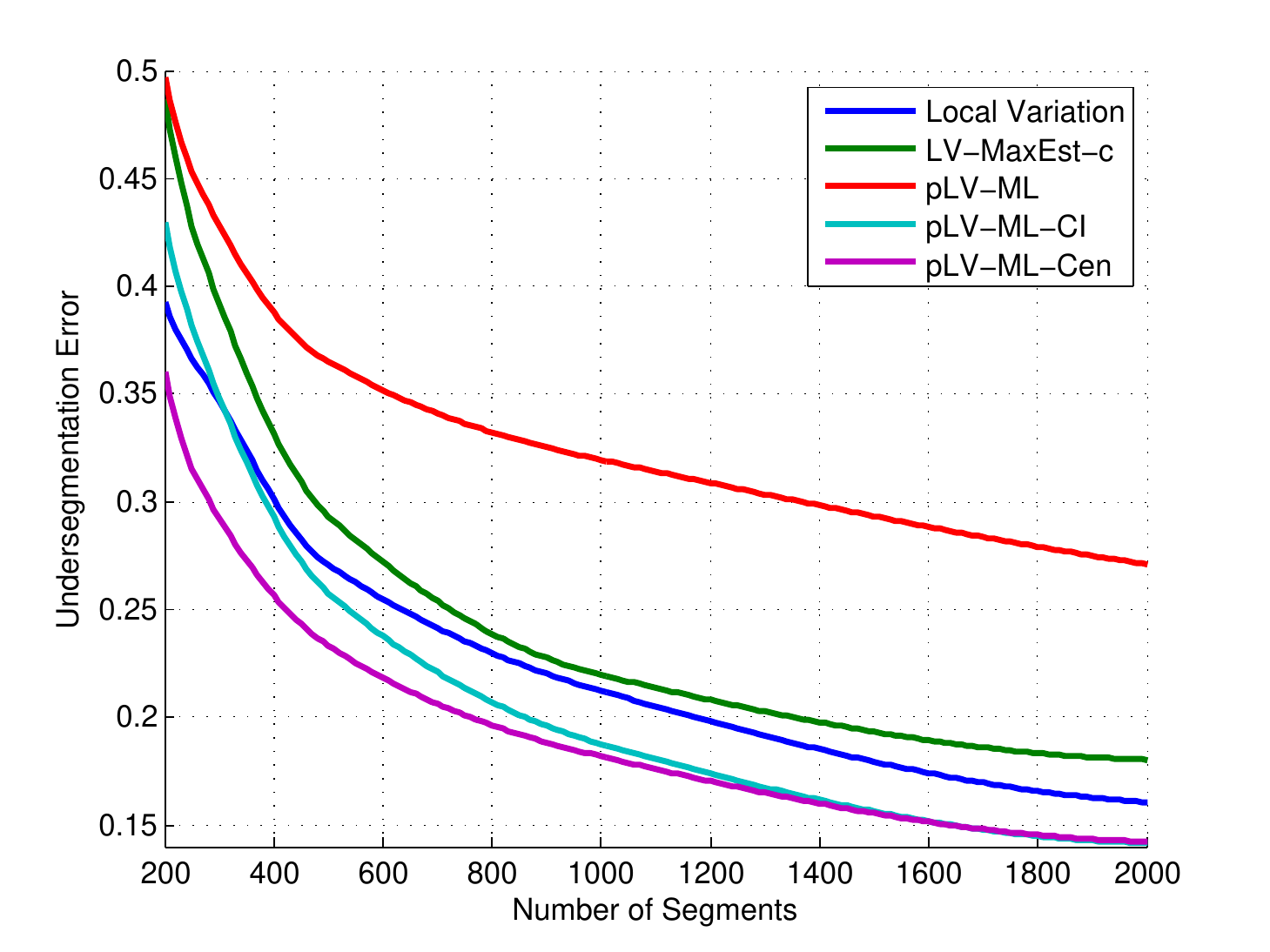}}
	\end{center}
	\caption{Recall (top) and undersegmentation error (bottom) of the LV method against its probabilistic versions proposed in this paper. The proposed hypothesis test (with censored estimation) performs best for medium to large segment counts.}
	\label{fig:HypothesisTestResults}
\end{figure}

The first probabilistic version, LV-MaxEst, is based on a uniform distribution model and estimates its interval. It is parameterless and gives a single segmentation (see figure \ref{fig:FHTests}). It is difficult to compare its average recall to that of other methods, because the segmentation of different images results in different numbers of segments (and different recall). Therefore, we use the generalized method, LV-MaxEst-c, for comparison; see section \ref{sec:UsingTheMaximumSampleStatistics}. The recall curve is shown in figure \ref{fig:HypothesisTestResults} (top). It is clearly inferior to that of LV but in a sense it behaves similarly and is better than several of the other methods in the literature.

The other probabilistic version relies on the exponential distribution model, which is specified by a single parameter $\lambda$. It is important to estimate this parameter carefully. The maximum likelihood approach, pLV-ML, overestimates $\lambda$, yielding poor recall; see figure \ref{fig:HypothesisTestResults} (top). Estimating the confidence interval (95\%) and using its lower limit as in pLV-ML-CI (section \ref{sec:HypothesisTesting}) results in the same recall as LV. Just like LV, the method depends on a single parameter $\delta$, determined by the number of segments selected.
Observing that the data is biased led to using censored estimation. The resulting algorithm (pLV-ML-Cen) gives a more accurate estimate of $\lambda$ and yields excellent, state-of-the-art, results.

\begin{figure}[tb]
	\begin{center}
		\subfigure{\includegraphics[width=0.23\textwidth]{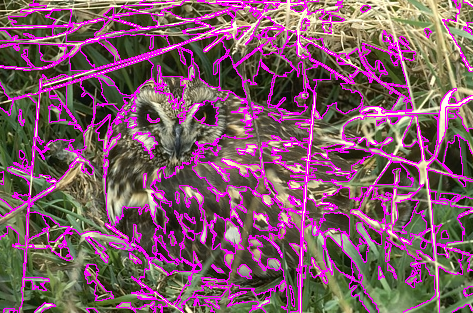}}
		\subfigure{\includegraphics[width=0.23\textwidth]{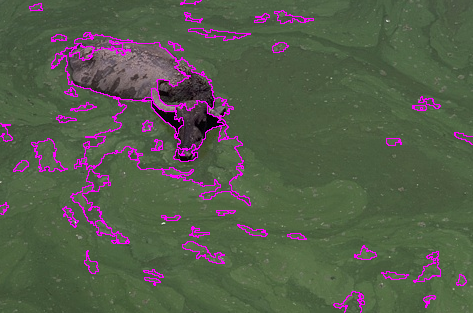}}
	\end{center}
	\caption{Two images segmented with the probabilistic local variation algorithm (censored version) with parameters $m=350,\delta = 0.05$. The number of segments depends on the characteristics of the image, yielding 268 (left) and 57 (right) segments respectively.}
	\label{fig:ExampleProbabilisticSegmentations}
\end{figure}

This algorithm depends on two parameters: $m$, the expected size of the segment, and $\delta$, the level of significance of the decision (see eq. \eqref{eq:RejectionRuleExpanded}). The number of segments is controlled by any combination of them.
For a prespecified number of segments $S$ (e.g., as needed for comparing different methods), we use the average segment size $m = ImageSize / S$, but found that setting $m$ to any large value (e.g., 200) works equally for $S \in [200,2000]$; $\delta$ is then tuned so that $S$ segments are obtained. This option was used for generating figure \ref{fig:HypothesisTestResults}.
Alternatively, $m,\delta$ can be set by empirically maximizing the average performance, which depends on the image set. Then, using these parameters for a specific image provides adaptive segmentation, yielding more segments on ``busy'' images and fewer segments on smooth ones.  See an example in figure \ref{fig:ExampleProbabilisticSegmentations}.
%
%
%
%

In terms of recall, probabilistic local variation improves LV and matches ERS to achieve the best results for large numbers of segments. The undersegmentation error is just behind that of ERS, best for this measure; see figure \ref{fig:AllMethodsResults}. The running time of our method is exactly the same as LV (0.3 sec on a Pentium 4GB machine), which is almost as fast as the fastest method (SLIC) and much faster than the only method that achieves the same recall (ERS, 2.5 sec).

\subsection{Multi-Class Segmentation Performance}
\label{sec:MultiClassSegmentationPerformance}
Extracting superpixels is not an end in itself, and therefore testing oversegmentation in the context of common tasks is important. Following the evaluation framework presented in \cite{SLICSuperpixels}, we examine the quality of pLV (specifically pLV-ML-Cen) in the context of the higher level task of multi-class segmentation. We perform our experiments on the MSRC 21-class database \citep{Msrc21Database} and use the segmentation method of \cite{GouldMultiClassSegmentation}, which proceeds as follows: the input image is divided into superpixels and a set of features is calculated on each superpixel, the calculated features are fed to a set of classifiers previously trained (one for each class), and the labeling is selected by minimizing an energy potential on a conditional random field. Table \ref{table:MulticlassSegmentation} presents the segmentation accuracy obtained when selecting different oversegmentation methods. Note that the accuracy achieved when using pLV is higher than when using the original LV algorithm. Furthermore, the accuracy obtained with pLV is improved only by the slower ERS algorithm.

\begin{table}[tb]
	\centering
	\tabcolsep=0.18cm
	\begin{tabular}{c c c c c c c}
	\hline
	pLV & LV & MS & TP & WS & SLIC & ERS \\
	77.0\% & 74.6\% & 70.3\% & 76.0\% & 74.7\% & 76.9\% &78.0\% \\
	\hline
	\end{tabular}
	\caption{Multiclass segmentation accuracy when using the following methods for the oversegmentation stage: pLV, LV, mean shift (MS), turbopixels (TP), watershed (WS), SLIC and ERS.}
	\label{table:MulticlassSegmentation}
\end{table}

\section{Discussion}
\label{sec:Discussion}

\subsection{Single-Linkage Clustering}
\label{sec:SingleLinkageClustering}
The LV algorithm, as well as its probabilistic versions considered here, are hierarchical algorithms, which, at every step, examine and  possibly merge the two most similar segments.

Being single-linkage algorithms, they specify the dissimilarity between two segments, $X$ and $Y,$ as the minimal distance between their two closest elements: $D(X,Y) = \underset{x \in X, y \in Y}{\min}d(x,y)$, where $d(x,y)$ is some underlying dissimilarity function between elements. Recall that in the graph based notation the dissimilarities between elements are represented by the corresponding edge weights and therefore $D(X,Y)$ is the minimal weight of an edge from $X$ to $Y$. In the LV and pLV case, the decision whether to merge two segments is made by testing, independently for each segment, whether the weight $D(X,Y)$ belongs to the distribution of weights estimated for the segment.

This paper focuses on the LV algorithm \citep{FelzSegmentation} and on the implied single-edge based merging decision.
Alternatively, the decision as to whether two segments should be merged can be made by testing whether all their weights may be explained by a single, common distribution.
For one-dimensional distributions characterizing a segment, classical tests such as the Kolmogorov-Smirnov test \citep{Ross2009Probability} may be used; see, e.g., \cite{PauwelsKolmogrovSmirnov}. See also \cite{PengAutomaticImageSegmentation} for another use of one\-/dimensional test for segmentation. Usually, however,  more effective, multidimensional distributions describing, say, the segment's texture or color, are preferred.
Formal tests for multidimensional distributions are problematic
(see, however, \cite{GLM:nips:12}). Usually the distribution is approximated as a mixture of visual words (textons) and the $\chi^2$ distance between the histograms serves as a measure of dissimilarity between segments. A more reliable dissimilarity measure can be obtained by augmenting this distance with edge information \citep{MartinLearningNaturalImageBoundaries}.

In the context of hierarchical segmentation, the distribution comparison technique may be used in two different ways. One approach carries out the merging process according to a fixed order determined before the process begins, as done in single-linkage processes.
This option is inconsistent with the hierarchical approach because the most similar pairs of segments (according to the merging criterion) are not tested before the others.
The other option, to recalculate the order dynamically after every merge,
is computationally expensive.

An effective combination studied in the literature is to divide the hierarchical merging process into stages. At the beginning of a stage, every segment is specified as an element, and the dissimilarity between the elements is calculated according to an arbitrary dissimilarity measure, which may depend on the distributions. Then all the merges in this stage proceed according to a single-linkage algorithm. This approach has shown to be a good trade-off between runtime and accuracy, as presented in \cite{KNKY:segmentation:11, RenISCRA, MB:thesis}.

\section{Conclusion}
\label{sec:Conclusion}
The local variation algorithm of \cite{FelzSegmentation} is a simple yet amazingly effective oversegmentation method. In this paper we analyzed the LV algorithm using statistical and empirical methods and showed that the algorithm and its performance may be explained by statistical principles.

We proposed an oversegmentation algorithm, denoted probabilistic local variation, that is based on hypothesis testing and on the statistical properties of natural images.

We found that probabilistic local variation is highly accurate, outperforms almost all other oversegmentation methods, and runs much faster than the one equally accurate oversegmentation algorithm (ERS). This is remarkable because it follows from a statistical interpretation of a 10-year old method (LV), which is, by the way, still one of the best competitors.

\begin{acknowledgements}

The authors would like to thank the US-Israel Binational Science Foundation, the Israeli Ministry of Science, Technology and Space, and the E. and J. Bishop Research Fund.

\end{acknowledgements}
\end{sloppypar}

\bibliographystyle{spbasic}      
\bibliography{Bibliography}   

\end{document}